\documentclass{article}
\usepackage{spconf,amsmath,graphicx,hyperref}
\usepackage{amssymb}
\usepackage{booktabs}
\usepackage{multirow}
\usepackage{algorithm}
\usepackage{algpseudocode}
\usepackage{cite}
\usepackage{balance}

\def\E{{\cal E}}
\def\a{{\text{a}}}
\def\v{{\text{v}}}
\def\p{{\text{p}}}

\title{Training-Free Multimodal Guidance for Video to Audio Generation}
\makeatletter
\def\@name{
\emph{Eleonora~Grassucci$^{1^*}$},
\emph{Giuliano Galadini$^{1,2^*}$},
\emph{Giordano Cicchetti$^{1^*}$},
\emph{Aurelio Uncini$^1$},\\
\emph{Fabio Antonacci$^2$},
\emph{Danilo Comminiello$^1$}

\thanks{This work was partially supported by the European Union under the NRRP of NextGenerationEU, partnership on “Future Artificial Intelligence Research” (PE00000013 – SPOKE 5 - CUP B53C22003980006 - FAIR: High Quality AI), and partially by the \textit{Progetti di Ateneo} of Sapienza University of Rome under grant RM123188F75F8072 and RM1241910FC4BEEA.}}
\address{$^1$Dept. of Information Engineering, Electronics, and Telecomm., Sapienza University of Rome, Italy\\$^2$Dept. of Electronics, Information, and Bioengineering, Politecnico di Milano, Italy\\ \small{$^*$Authors contributed equally}}
\begin{document}
%\ninept
%
\maketitle
\ninept
\begin{abstract}
Video-to-audio (V2A) generation aims to synthesize realistic and semantically aligned audio from silent videos, with potential applications in video editing, Foley sound design, and assistive multimedia. Although the excellent results, existing approaches either require costly joint training on large-scale paired datasets or rely on pairwise similarities that may fail to capture global multimodal coherence. In this work, we propose a novel training-free multimodal guidance mechanism for V2A diffusion that leverages the volume spanned by the modality embeddings to enforce unified alignment across video, audio, and text.
% At each denoising step, the intermediate clean audio latent is projected into a shared embedding space along with video and text embeddings, and its geometric volume is minimized to ensure semantic consistency.
The proposed multimodal diffusion guidance (MDG) provides a lightweight, plug-and-play control signal that can be applied on top of any pretrained audio diffusion model without retraining. Experiments on VGGSound and AudioCaps demonstrate that our MDG consistently improves perceptual quality and multimodal alignment compared to baselines, proving the effectiveness of a joint multimodal guidance for V2A.
% These results confirm the effectiveness of GRAM-guided diffusion as a robust and efficient strategy for video-to-audio generation.
\end{abstract}


\section{Introduction}
\label{sec:intro}

The ability to generate realistic and semantically consistent audio from a given video has the potential to transform a wide range of applications, from automated video editing and silent film restoration to assistive multimedia technologies and immersive content creation \cite{Comunit2023SyncfusionMO,Liu2023AudioLDM2L,MarinoniIJCNN2025}. The task, known as video-to-audio (V2A) generation, requires not only synthesizing plausible acoustic content, but also ensuring that the generated audio aligns semantically with the visual input and, potentially, with an additional textual description. 

Recent approaches to V2A generation have made impressive progress by combining pretrained audio and video backbones with several conditional generation strategies \cite{Comunit2023SyncfusionMO, Cheng_2025_CVPR, gramaccioni2024stablev2a, Jeong_Kim_Chun_Lee_2025, yang2024drawaudioleveragingmultiinstruction, GramaccioniIJCNN2025}. These methods often rely on diffusion models to synthesize high-fidelity waveforms or spectrograms, guided by representations extracted from the visual domain or from diverse sources like text. To improve semantic grounding, many incorporate auxiliary textual prompts or leverage joint training across multiple modalities \cite{Kushwaha_2025_CVPR, Li_Yang_Mao_Ye_Chen_Zhong_2025, gramaccioni2024stablev2a}.
% qui problemi dei metodi attuali:
% - spesso richiedono un joint training, paired data, long time
% - high computational requirements
% - low one (seeing and hearing) proposes to guide the generation via imagebind, which does not produce meaningful multimodal latent space
However, existing approaches may suffer from several key limitations. By requiring joint training of video and audio models, the computational demand is usually very high, often needing days to train on multiple GPUs \cite{NEURIPS2024_e7384de3, Cheng_2025_CVPR}. Furthermore, these systems usually also require large-scale paired datasets \cite{VGGSOUND}, which are expensive to collect and rare to find. This limits all the models to be trained on the same training datasets, which have to be cleaned and carefully preprocessed to remove unmeaningful audio-video pairs \cite{gramaccioni2024stablev2a, Cheng_2025_CVPR}.
In response to these challenges, recent work has explored training-free paradigms, such as Seeing\&Hearing \cite{Xing2024SeeingAH}, which uses the pretrained ImageBind \cite{Girdhar2023ImageBindOE} to guide audio generation from visual inputs. Yet, ImageBind only aligns modalities via pairwise cosine similarity and may not construct a coherent joint embedding space across video, audio, and text in practice \cite{jeong2025anchorsaweighsailoptimal, cicchetti2025gram}. As a result, the guidance signal may lack geometric consistency and may lead to semantically misaligned generations.

\begin{figure}
    \centering
    \includegraphics[width=\linewidth]{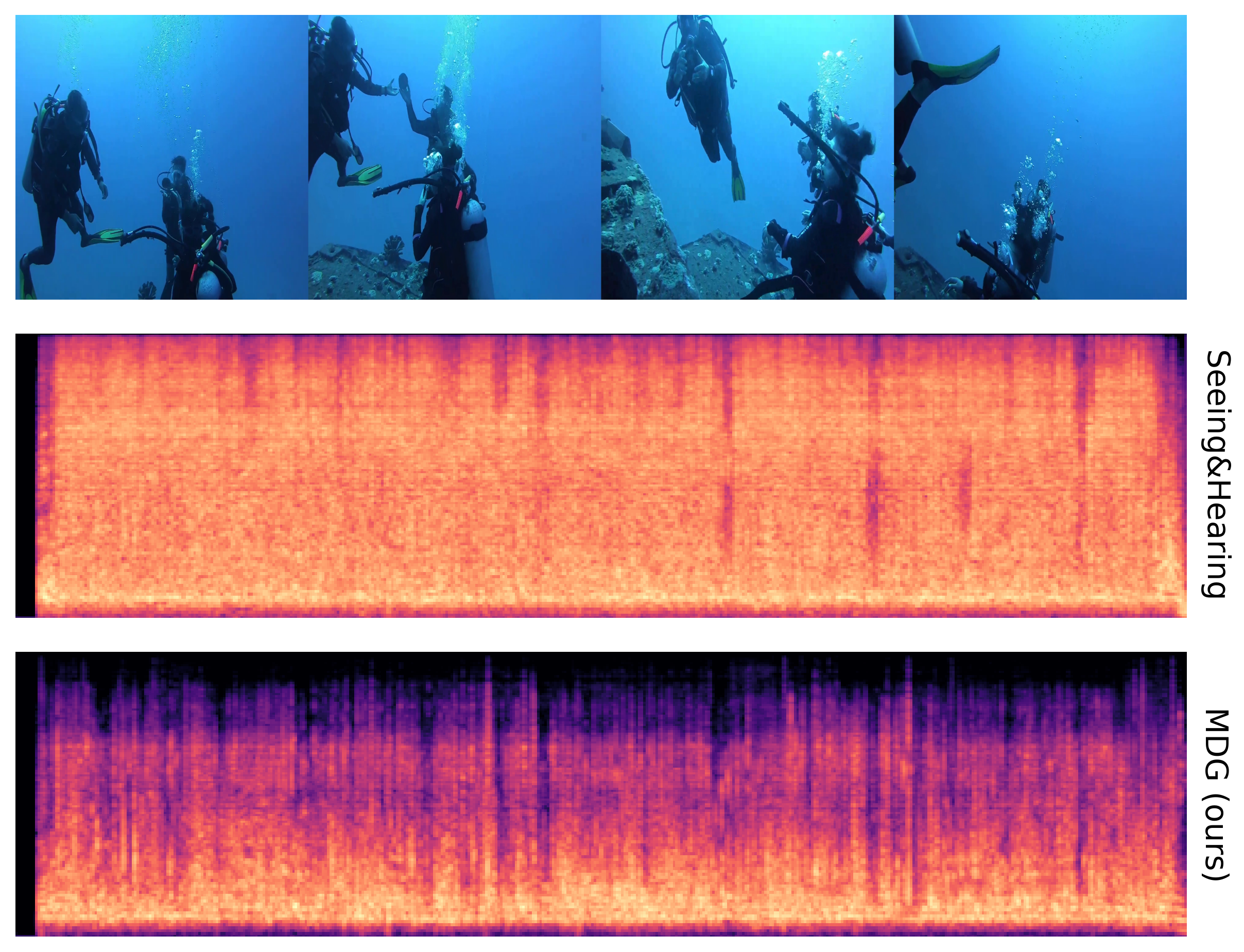}
    \caption{Generated samples from VGGSound by Seeing\&Hearing \cite{Xing2024SeeingAH} and by our method. MDG (ours) better guides the generation towards a semantically meaningful synthesized audio, while Seeing\&Hearing generation is noisy and semantically inconsistent.}
    \label{fig:results}
\end{figure}

In this work, we introduce a novel multimodal guidance mechanism for V2A diffusion models that translates geometric and unified multimodal consistency into a control signal for generation. Specifically, we show that the volume spanned by audio-video-text embeddings in a shared latent space can be used to steer the audio denoising process toward semantic consistency. Intuitively, when the three modalities refer to the same underlying semantic content, their embeddings have a low-volume configuration in the shared space. By minimizing this volume during denoising, we encourage the generated audio to converge toward a representation that is jointly compatible with both the video and the textual prompt. The proposed multimodal diffusion guidance (MDG) operates in a training-free, plug-and-play fashion: we inject the unified volume-based alignment measure into the latent denoising loop of a pretrained audio diffusion model without requiring any retraining of the generative backbone or the encoders.
% At each timestep, we compute a clean audio latent prediction, evaluate its alignment with the video and text conditions, and adjust the latent accordingly to reduce the geometric volume across modalities.
% This strategy generalizes existing classifier-free or embedding-guided diffusion mechanisms, while capturing richer multi-way semantic relationships.
We evaluate our approach on well-known benchmarks such as VGGSound and AudioCaps, comparing it against both traditional V2A baselines and recent latent aligner models. Results show that our method significantly improves semantic consistency and perceptual quality of generated audio, while remaining efficient.\\Our main contributions can be summarized as follows:
\begin{itemize}
  \item We introduce a novel training-free sampling strategy with latent multimodal guidance requiring no paired large training data, no joint training, and no modification to the diffusion backbone.
  \item We propose a multimodal joint guidance mechanism for video-to-audio generation, which steers a pretrained diffusion model using a volume-based tri-modal alignment objective across video, audio, and text.
  \item We show that our volume-guided inference outperforms prior cosine-based aligners by promoting stronger multimodal consistency and reducing semantic drift in the generated audio.
\end{itemize}

\section{Proposed Method}
\label{sec:method}

\begin{figure}
    \centering
    \includegraphics[width=\linewidth]{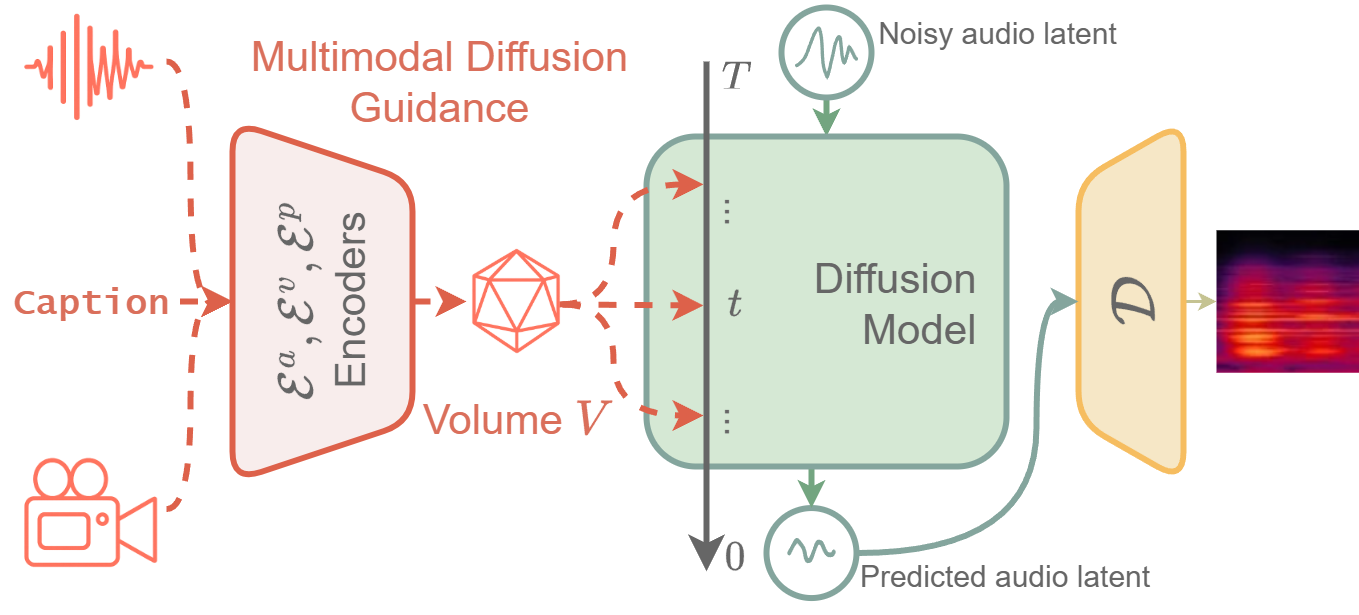}
    \caption{Overview of the proposed multimodal diffusion guidance (MDG) for the generation process.}
    \label{fig:method}
\end{figure}

We consider the task of generating audio $\hat{\mathbf{x}}^\a$ conditioned on a given video $\mathbf{x}^\v$ and a text prompt $\mathbf{x}^\p$. The audio generator is a latent diffusion model (LDM) operating in the audio latent space. By jointly leveraging the information of the clean audio $\mathbf{x}^\a$, of the video $\mathbf{x}^\v$ (and optionally of the text $\mathbf{x}^\p$), we provide a novel unified geometric guidance signal to the generative model. Unlike prior latent aligners based on pairwise cosine similarity like ImageBind \cite{Girdhar2023ImageBindOE}, we enforce joint consistency across multiple modalities (video, audio, text) through a volume-based objective, which better exploits the cross-modal information of the multimodal data. Figure~\ref{fig:method} provides an overview of the proposed multimodal diffusion guidance (MDG), with the multimodal guidance brought by the volume signal that guides the generation process of the audio latent.

\subsection{Latent Diffusion for Audio}
Let $\mathcal{E}, \mathcal{D}$ be the generic audio autoencoder and $\mathbf{z}^\a_0=\mathcal{E}(\mathbf{x}^\a)$ the clean audio latent. The forward process is

\begin{align}
q(\mathbf{z}^\a_t \mid \mathbf{z}^\a_{t-1})=\mathcal{N}\!\big(\sqrt{1-\beta_t}\,\mathbf{z}^\a_{t-1},\,\beta_t \mathbf{I}\big), \\
q(\mathbf{z}^\a_t \mid \mathbf{z}^\a_0)=\mathcal{N}\!\big(\sqrt{\bar\alpha_t}\,\mathbf{z}^\a_0,\,(1-\bar\alpha_t)\mathbf{I}\big),
\label{eq:forward}
\end{align}

\noindent with $\bar\alpha_t=\prod_{i=1}^t (1-\beta_i)$. The denoiser $\epsilon_\theta^\a(\mathbf{z}^\a_t,t,\mathbf{x}^\p)$ is pretrained by

\begin{equation}
\mathcal{L}_{\text{LDM}}=\mathbb{E}_{\mathbf{z}^\a_0,\epsilon,t}\left\|\epsilon-\epsilon_\theta^\a(\mathbf{z}^a_t,t,\mathbf{x}^\p)\right\|_2^2.
\label{eq:ldm}
\end{equation}

\noindent During sampling, the predicted clean latent is

\begin{equation}
\tilde{\mathbf{z}}^\a_0=\frac{1}{\sqrt{\bar\alpha_t}}\Big(\mathbf{z}^\a_t-\sqrt{1-\bar\alpha_t}\,\epsilon_\theta^\a(\mathbf{z}^\a_t,t,\mathbf{x}^\p)\Big),
\label{eq:clean-pred}
\end{equation}

\noindent which we will steer with the proposed guidance. The final spectrogram is $\hat{\mathbf{x}}^\a=\mathcal{D}(\tilde{\mathbf{z}}^\a_0)$ that can be converted into a waveform with any vocoder. As a latent audio diffusion model backbone, and respective encoder, decoder, and vocoder, in our work, we employ the well-known AudioLDM \cite{Liu2023AudioLDMTG}.

\subsection{Multimodal Contrastive Loss Function}
Let $\E^\v,\E^\a,\E^\p$ be GRAM \cite{cicchetti2025gram} encoders mapping video frames, audio spectrogram, and text into a shared $D$-dimensional space:

\begin{equation}
\mathbf{e}^\v=\E^\v(\mathbf{x}^\v),\quad
\mathbf{e}^\a=\E^\a(\mathbf{x}^\a),\quad
\mathbf{e}^\p=\E^\p(\mathbf{x}^\p), \quad \in\mathbb{R}^D
\label{eq:mod_emb}
\end{equation}

\noindent with $\|\mathbf{e}^m\|_2=1$, $m=\{\v,\a,\p\}$. The resulting embeddings can be arranged in the matrix $\mathbf{Z}=[\mathbf{e}^\v,\,\mathbf{e}^\a,\,\mathbf{e}^\p]\in\mathbb{R}^{D\times 3}$. According to \cite{Gantmacher1959matrix}, from $\mathbf{Z}$ it is possible to build the Gram matrix $\mathbf{K} \in \mathbb{R}^{3 \times 3}$defined as:

\begin{equation}
\mathbf{K} \;=\; \mathbf{Z}^\top \mathbf{Z},
\label{eq:gram}
\end{equation}

\noindent and define the volume of the high-dimensional parallelotope spanned by modality embeddings $\mathbf{e}^\v,\,\mathbf{e}^\a,\,\mathbf{e}^\p$, which is:

\begin{equation}
V \;=\; \sqrt{\det \mathbf{K}}.
\label{eq:volume}
\end{equation}

Intuitively, the volume of the parallelotope spanned by the matching triplets $(\mathbf{e}^\v_i,\mathbf{e}^\a_i,\mathbf{e}^\p_i)$ should be small as they share the same semantics. On the contrary, the volume of non-matching triplets $(\mathbf{e}^\v_i,\mathbf{e}^\a_j,\mathbf{e}^\p_k)$, with $i\neq j \neq k$ should be larger, representing dissimilar semantic concept. Therefore, the volume $V$ can be involved as a semantic similarity measure in place of the pairwise cosine similarity in the contrastive loss \cite{cicchetti2025gram} as:

\begin{align}
\label{eq:contrastiveloss}
    \mathcal{L}_{\text{AV2T}}&=-\frac{1}{B}\sum_{i=1}^{B}\log\frac{\exp(-V(\mathbf{e}^\v_{i},\mathbf{e}^\a_{i},\mathbf{e}^\p_i)/\tau)}{\sum_{j=1}^{K}\exp(-V(\mathbf{e}^\v_{i},\mathbf{e}^\a_{i},\mathbf{e}^\p_j)/\tau)},\\
    \mathcal{L}_{\text{T2AV}}&=-\frac{1}{B}\sum_{i=1}^{B}\log\frac{\exp(-V(\mathbf{e}^\v_{i},\mathbf{e}^\a_{i},\mathbf{e}^\p_i)/\tau)}{\sum_{j=1}^{K}\exp(-V(\mathbf{e}^\v_{j},\mathbf{e}^\a_{j},\mathbf{e}^\p_i)/\tau)},
\end{align}

\noindent with temperature $\tau>0$. By design, Eq.~\eqref{eq:contrastiveloss} is the standard InfoNCE \cite{Oord2018RepresentationLW} with pairwise cosine similarity replaced by the volume measure of Eq.~\eqref{eq:volume}. This retains the benefits of contrastive learning while enforcing joint (tri-modal) geometry rather than pairwise alignment.
% The final contrastive loss, equal to the average of the two non-symmetric losses, is formally:
% \begin{equation}
% \mathcal{L}_{\text{contr}} = \frac{1}{2} (\mathcal{L}_{AV2T} + \mathcal{L}_{T2AV}).
% \end{equation}

\subsection{Multimodal Diffusion Guidance}

We propose a training-free multimodal diffusion guidance (MDG) mechanism that leverages the geometric structure learned by GRAM \cite{cicchetti2025gram} to steer the audio generation process. At each denoising step, the model predicts an intermediate clean audio latent, which is then adjusted through the shared information coming from the multimodal embedding space built with the video and text inputs. Rather than relying on pairwise similarities as \cite{Xing2024SeeingAH}, we compute a tri-modal similarity measure based on the volume spanned by the three embeddings. This allows the generative model to be guided 
% This measure serves as a semantic alignment signal, indicating how well the current generation matches the intended video and text conditioning.
We compute the audio latent $\tilde{\mathbf{z}}^\a_0$ with Eq.~\eqref{eq:clean-pred} and the embeddings $\mathbf{e}^\v_i,\mathbf{e}^\a_i,\mathbf{e}^\p_i$ with Eq.~\eqref{eq:mod_emb}, then we leverage the embeddings to guide the diffusion step through:

\begin{equation}
\mathbf{z}_t^\a \;\leftarrow\; \mathbf{z}_t^\a \;-\; \eta\, \nabla_{\mathbf{z}_t^\a}V,
\label{eq:update}
\end{equation}

\noindent with $\eta$ learning rate and $V$ the volume computation in Eq.~\eqref{eq:volume}. In this way, we can iteratively improve the alignment between the evolving audio latent and the reference modalities by adjusting the latent to reduce the disagreement in the joint embedding space. In our formulation, this disagreement is quantified by the geometric volume formed by the embeddings, which shrinks when the three modalities represent the same semantic content and grows otherwise. By repeatedly reducing this volume over successive denoising steps, the model gradually enforces semantic consistency in the generated audio with respect to the visual and textual conditions. Since the visual content remains fixed throughout the generation, the guidance focuses solely on refining the audio trajectory. The procedure is inherently modular and lightweight, as it does not modify the underlying diffusion model and can be applied on top of any pretrained audio generator without requiring paired training data. In detail, each guidance step adds $O(D)$ for embeddings and $O(1)$ for $3\times3$ volume computation. Therefore, the method remains training-free and lightweight.
The proposed sampling algorithm for multimodal diffusion guidance is described in Algorithm~\ref{alg:gram-v2a}.

\begin{algorithm}[t]
\caption{Multimodal-guided V2A sampling}
\label{alg:gram-v2a}
\begin{algorithmic}[1]
\Require Learning rate $\eta$; optimization steps $N$; $K$ warmup steps; prompt $\mathbf{x}^\p$
\State $\mathbf{e}^\p \gets \E^\p(\mathbf{x}^\p)$
\State $\mathbf{e}^\v \gets \E^\v(\mathbf{x}^\v)$
\For{$t = T$ \textbf{to} $0$}
  % \State $\mathbf{z}^{\a}_{t} \gets \text{DENOISE}(\mathbf{z}^{\a}_{t+1}, \mathbf{e}^\p)$
  \State $\mathbf{z}^{\a}_{t} \gets \epsilon_\theta^\a(\mathbf{z}^\a_{t+1},t,\mathbf{e}^\p)$
  \If{$t < K$}
    \For{$n = 0$ \textbf{to} $N$}
      \State $\tilde{\mathbf{z}}^{\a}_{0} \gets \dfrac{1}{\sqrt{\bar{\alpha}^{\a}_{t}}}
      \Big( z^{\a}_{t} - \sqrt{1-\bar{\alpha}^{\a}_{t}}\, \epsilon^{\a}_{t} \Big)$
      \State $\mathbf{e}^{\a} \gets \E^\a(\mathbf{z}^{\a}_{0})$
      \State $\mathbf{K} \gets (\mathbf{e}^{\v},\mathbf{e}^{\a},\mathbf{e}^{\p})$
      \State $V \gets \sqrt{\det(\mathbf{K})}$
      \State $\hat{\mathbf{z}}^{\a}_{t} \gets \mathbf{z}^{\a}_{t} - \eta\nabla_{z^{\a}_{t}}V$
      % \State $\hat{\mathbf{e}}^p \gets \mathbf{e}^p - \lambda_{2}\nabla_{\mathbf{e}^p}\mathcal{L}_{\text{contr}}$
    \EndFor
  \EndIf
\EndFor
\State \Return $\mathbf{z}^{\a}_{0}$
\end{algorithmic}
\end{algorithm}

\section{Experiments}
\label{sec:exp}
This section details the experimental framework designed to evaluate the performance of the proposed Multimodal Diffusion Guidance.
% We describe the experimental and models setup, define the evaluation metrics used to measure audio quality and audio-visual alignment, and present a quantitative analysis of the results.

\subsection{Experimental Setup}
To conduct a comprehensive comparison, we design a two-part experimental evaluation to assess both in-domain performance and generalization capabilities under a domain shift.
The main experiments are performed on the VGGSound dataset~\cite{VGGSOUND}, a large-scale collection of video clips focused on specific sound events. This dataset is the primary benchmark in the Seeing\&Hearing paper \cite{Xing2024SeeingAH}, to which we are particularly interested in a comparison being their pairwise-based diffusion guidance. Since the proposed MDG method does not require training, we consider only the test set of VGGSound, and we run the evaluation on 3k samples from the test set, following \cite{Xing2024SeeingAH}.
Furthermore, we perform an off-domain experiment on the AudioCaps dataset~\cite{AUDIOCAPS}, being the backbone AudioLDM pretrained on VGGSound. Unlike the event-centric nature of VGGSound, AudioCaps audio tracks may be less directly correlated with the visual content, as the sound source may not be visible in the video. This provides a challenging test for generalization. Out of the official test set, we find that only 697 videos contained usable audio, and our evaluation is performed on this subset. As comparisons, we run experiments with two well-established frameworks, namely SpecVQGAN \cite{SpecVQGAN_Iashin_2021} and Diff-Foley \cite{NEURIPS2023_98c50f47} on VGGSound. Furthermore, for a more detailed evaluation, our primary comparison is Seeing\&Hearing framework~\cite{Xing2024SeeingAH}, which, similar to our method, employs a training-free, diffusion-guided approach. However, Seeing\&Hearing leverages ImageBind \cite{Girdhar2023ImageBindOE} for the guidance, which is based on pairwise alignment between modalities. The main scope of our experimental evaluation is to demonstrate the superiority of the proposed multimodal guidance based on the tri-modal volume over the pairwise-based guidance of Seeing\&Hearing.

\subsection{Implementation Details}
We conduct all experiments on a single NVIDIA Quadro RTX 8000 with 48GB of VRAM. The audio generation process employs a pre-trained AudioLDM model from the cvssp/audioldm-m-full checkpoint~\footnote{https://huggingface.co/cvssp/audioldm-m-full}. For guidance, the Seeing\&Hearing baseline leverages the pre-trained ImageBind-Huge model, while, for our MDG, we utilize the GRAM encoders (EVAClip-ViT-G for video, BEATS for audio, and BERT-B for text) \cite{cicchetti2025gram} from the GRAM\_pretrained\_4modalities checkpoint~\footnote{https://github.com/ispamm/GRAM}. For a fair comparison, all hyperparameters are adopted from the official Seeing\&Hearing repository~\footnote{https://github.com/yzxing87/Seeing-and-Hearing}. The denoising process runs for 30 DDIM steps with a guidance scale of 2.5. The external guidance is activated after the first 20\% of the steps, updating the latent variable at each subsequent step with a single optimization step using an Adam optimizer and a learning rate $\eta$ of 0.1. To process the guidance modality, we sample 2 video frames evenly across the entire video duration, and 1 audio chunk for a 10-second frame, following \cite{cicchetti2025gram}. 

None of our experiments requires any kind of training or large-scale training dataset, as the proposed method is training-free and can be employed on top of any pretrained audio latent diffusion model to guide the generation process.

\begin{table*}
\centering
\caption{Quantitative results of generated audio samples.}
\label{tab_results}
\begin{tabular}{clcccccc}
\toprule
Dataset & Method & FAD $\downarrow$ & FAVD $\downarrow$ & PEAVS $\uparrow$ & KL $\downarrow$ & ISc $\uparrow$ & FD $\downarrow$ \\
\midrule
\multirow{4}{*}{VGGSound} & SpecVQGAN \cite{SpecVQGAN_Iashin_2021} & 7.74 & - & - & 3.29 & 5.11 & 37.27 \\
& Diff-Foley \cite{NEURIPS2023_98c50f47} & 8.91 & 3.57 & 3.15 & 3.31 & 4.28 & 38.11 \\
& Seeing\&Hearing \cite{Xing2024SeeingAH} & 7.80 & 3.44 & 2.90 & 3.35 & 4.88 & 37.68 \\
& Ours      & \textbf{6.04 }& \textbf{2.60} & \textbf{3.40} & \textbf{2.78} & \textbf{5.88} & \textbf{31.95} \\ \midrule
\multirow{2}{*}{AudioCaps} & Seeing\&Hearing \cite{Xing2024SeeingAH} & 11.04 & 4.44 & 3.02 & 3.43 & \textbf{4.68} & 51.92 \\
& Ours      & \textbf{10.77} & \textbf{4.31} & \textbf{3.07} & \textbf{3.40} & \textbf{4.68} & \textbf{51.05} \\
\bottomrule
\end{tabular}
\end{table*}

\subsection{Evaluation Metrics}
To assess the performance of evaluated methods, we evaluate both the perceptual quality of the generated audio and its alignment with the source video. We employ a suite of objective metrics from the open-source AVGen-Eval Toolkit provided by Amazon Science~\cite{goncalves2024peavsperceptualevaluationaudiovisual}. For audio quality, we use the Fréchet Audio Distance (FAD)~\cite{kilgour2019frechetaudiodistancemetric}, Fréchet Distance (FD), Kullback-Leibler (KL) Divergence, and the Inception Score (ISc)~\cite{salimans2016improvedtechniquestraininggans} to measure the fidelity, diversity, and statistical similarity of the generated audio to real-world samples. For audio-visual coherence, we measure the Fréchet Audio-Visual Distance (FAVD) to evaluate the joint distribution of the audio and video embeddings, and the Perceptual Evaluator of Audio-Visual Synchronization (PEAVS) \cite{goncalves2024peavsperceptualevaluationaudiovisual}, a learned metric designed to predict the temporal alignment of audio-visual events.

\subsection{Qualitative Analysis}
Figure~\ref{fig:results} presents a comparison between the spectrogram generated by Seeing\&Hearing and the proposed method on a random sample from the VGGSound test set. The perceptual evaluation reveals a stark difference in quality. The audio generated by the proposed MDG appears much better, presenting a soundscape appropriate for an underwater environment. In contrast, the audio from the baseline method degenerates into broadband, undifferentiated noise. Its spectrogram is largely uniform and lacks discernible features, which corresponds to the perceptually noisy and unstructured audio. This suggests that the pairwise alignment struggles with the limited global information, collapsing to a generic, poorly aligned output, whereas the proposed approach successfully captures the essential audio-visual correspondence.

\begin{table}
\centering
\caption{Semantic consistency results of generated audio with caption and video from the VGGSound test set. Comparison between the pairwise-guided Seeing\&Hearing and the proposed MDG.}
\resizebox{\linewidth}{!}{
\begin{tabular}{lccccc}
\toprule
Method & $V \downarrow$ & $\delta_{\text{cos}} \downarrow$ & $\delta_{\text{cos}}^{t,v} \downarrow$ & $\delta_{\text{cos}}^{t,a} \downarrow$ & $\delta_{\text{cos}}^{v,a} \downarrow$ \\
\midrule
Seeing\&Hearing \cite{Xing2024SeeingAH} & 0.937 & 2.488 & 0.703 & 0.891 & 0.893 \\
MDG (ours)      & \textbf{0.819} & \textbf{2.068} &  \textbf{0.517} & \textbf{0.713} & \textbf{0.838} \\
\bottomrule
\end{tabular}}
\label{tab:semantic}
\end{table}

\subsection{Quantitative Analysis}
Table~\ref{tab_results} presents the quantitative analysis results on the audio generation quality. On the in-domain VGGSound dataset, our MDG demonstrates a significant improvement over both baselines across all evaluation metrics. The proposed guidance achieves an FAD score of 6.04, substantially outperforming both Seeing\&Hearing (7.80) and SpecVQGAN (7.74), which indicates a marked increase in the perceptual quality of the generated audio. More importantly, our method obtains an FAVD of 2.60 and a PEAVS score of 3.40, showing a clear enhancement in audio-visual coherence compared to the pairwise-guided Seeing\&Hearing. To assess generalization capabilities, we further evaluate the models on the AudioCaps dataset, which represents a significant domain shift. On this more challenging task, our method continues to outperform Seeing\&Hearing on nearly every metric. We achieve a lower FAD and FAVD, and a higher PEAVS score, demonstrating that the unified multimodal alignment provided by the volume minimization is more robust and generalizes more effectively than the pairwise approach, even when faced with out-of-domain data.

\subsection{Semantic Consistency Analysis}

Upon the generated audio quality evaluation, we further analyze the semantic consistency of the generated audio to prove the effectiveness of the proposed multimodal diffusion guidance with respect to the pairwise baseline. We evaluate how close the semantics of the generated audio are to the original video and the descriptive caption with diverse metrics. First, we compute the volume $V$ in Eq.~\eqref{eq:volume} among the three embeddings. Second, we compute all the cross-modal cosine similarities as:

\begin{equation}
    \delta_{\text{cos}}^{m,n} = 1 - \cos(\mathbf{z}^m, \mathbf{z}^n),
\end{equation}

\noindent with $m,n=\{\v,\a\,\p\}$ being two modalities among video, audio and text prompts. For all the metrics, the lower the better, representing a smaller distance among embeddings and thus more similar semantic content. Table~\ref{tab:semantic} reports the results. The proposed MDG generates audio that is semantically more aligned with the original video and its caption according to all the metrics considered. These results prove that the proposed method effectively guides the generation process towards a more semantically aligned generation process, generating audio of better quality and with consistent semantics with the provided video and description.

\section{Conclusion}
\label{sec:con}
In this paper, we presented a novel multimodal guidance mechanism (MDG) for video-to-audio (V2A) generation that leverages the volume spanned by modality embeddings to enforce joint semantic alignment across video, audio, and text. MDG introduces a training-free, plug-and-play strategy that integrates a volume-based multimodal alignment objective into the denoising process of a pretrained audio diffusion model. Unlike prior approaches that rely on pairwise similarities or require expensive joint training, our solution provides a lightweight and effective control signal during inference. Experimental results on both VGGSound and AudioCaps demonstrate that our approach significantly improves audio quality and multimodal semantic consistency compared to baseline methods such as ImageBind-guided generation. These results confirm the effectiveness of leveraging an effective multimodal structure for controlling generative diffusion models, without modifying the underlying architecture or requiring paired training data.

\ninept
\balance
\bibliographystyle{IEEEbib}
\bibliography{strings,refs}

\end{document}